\documentclass[sigconf]{acmart}

\usepackage{graphicx}
\usepackage{comment}
\usepackage{amsmath} 
\usepackage{color}
\usepackage{multirow}

\usepackage{url}
\usepackage{algorithm}
\usepackage{algorithmic}
\usepackage{gnuplottex}
\usepackage{caption}
\usepackage{subcaption}
\usepackage{flushend}

\usepackage{xspace}
\newcommand{\sys}{\texttt{SMTM}\xspace}

\usepackage{hyperref}
\usepackage[utf8]{inputenc}

\AtBeginDocument{%
  \providecommand\BibTeX{{%
    \normalfont B\kern-0.5em{\scshape i\kern-0.25em b}\kern-0.8em\TeX}}}

\setcopyright{acmcopyright}

\begin{document}

\title{Boosting Mobile CNN Inference through Semantic Memory}

\fancyhead{}

\author{Yun Li}
\authornote{Contribution during internship at Microsoft Research}
\affiliation{%
	\institution{University of Science and Technology of China}
	\city{}
	\country{}
}
\email{yli001@mail.ustc.edu.cn}

\author{Chen Zhang}
\authornote{Corresponding author}
\affiliation{%
	\institution{Damo Academy, Alibaba Group}
	\city{}
	\country{}
}
\email{mingchong.zc@alibaba-inc.com}

\author{Shihao Han}
\authornotemark[1]
\affiliation{%
	\institution{Rose-Hulman Institute of Technology}
	\city{}
	\country{}
}
\email{hans3@rose-hulman.edu}

\author{Li Lyna Zhang}
\affiliation{%
	\institution{Microsoft Research}
	\city{}
	\country{}
}
\email{lzhani@microsoft.com}

\author{Baoqun Yin}
\authornotemark[2]
\affiliation{%
	\institution{University of Science and Technology of China}
	\city{}
	\country{}
}
\email{bqyin@ustc.edu.cn}

\author{Yunxin Liu}
\affiliation{%
	\institution{Institute for AI Industry Research (AIR), Tsinghua University}
	\city{}
	\country{}
}
\email{liuyunxin@air.tsinghua.edu.cn}

\author{Mengwei Xu}
\affiliation{%
	\institution{State Key Laboratory of Networking and Switching Technology, Beijing University of Posts and Telecommunications}
	\city{}
	\country{}
}
\email{mwx@bupt.edu.cn}

\begin{abstract}
Human brains are known to be capable of speeding up visual recognition of repeatedly presented objects through faster memory encoding and accessing procedures on activated neurons.
For the first time, we borrow and distill such a capability into a semantic memory design, namely \sys, to improve on-device CNN inference.
\sys employs a hierarchical memory architecture to leverage the long-tail distribution of objects of interest, and further incorporates several novel techniques to put it into effects:
(1) it encodes high-dimensional feature maps into low-dimensional, semantic vectors for low-cost yet accurate cache and lookup;
(2) it uses a novel metric in determining the exit timing considering different layers' inherent characteristics;
(3) it adaptively adjusts the cache size and semantic vectors to fit the scene dynamics.
\sys is prototyped on commodity CNN engine and runs on both mobile CPU and GPU.
Extensive experiments on large-scale datasets and models show that \sys can significantly speed up the model inference over standard approach (up to 2$\times$) and prior cache designs (up to 1.5$\times$), with acceptable accuracy loss.
\end{abstract}




\settopmatter{printacmref=false,printccs=false, printfolios=true} 
\renewcommand\footnotetextcopyrightpermission[1]{} 
\pagestyle{plain} 

\maketitle

\section{Introduction}

The recent advances of Convolutional Neural Networks (CNNs) have catalyzed many emerging mobile vision tasks, including but not limited to augmented reality, face recognition, activity recognition, etc~\cite{yi2020eagleeye,zeng2017mobiledeeppill,hodges2006sensecam}.
A notable trend is on-device CNN inference as against cloud offloading due to the tight delay constraint and data privacy concerns~\cite{gdpr}.
For instance, the Android applications empowered by on-device deep learning have increased by 27\% within only a quarter in 2018, where CNNs dominate the use cases ($>$85\%) \cite{xu2019first}.

The key challenge to fit CNN to resource-constrained mobile devices is its high computation load, especially in continuous vision tasks where predictions are performed on a stream of image frames.
A unique opportunity to accelerate continuous vision inference resides in its high temporal locality: recently seen objects are more likely to appear in the next few frames, and the frequency of object occurrence in the vision streams typically follows a long-tail distribution.
Those observations straightforwardly motivate a CNN system to ``memorize'' the recent inference results and directly omit the future inference if similar inputs are observed.

Indeed, such a memory mechanism also exists in human brains, from which neural network borrows its spirits exactly.
Human brain leverages temporal redundancy with \textit{priming effect}, a psychology phenomenon whereby exposure to one stimulus improves a response to a subsequent stimulus, without conscious guidance or intention~\cite{primed,priming}. Dated back to the 70s, biologic experiments~\cite{scarb1979} already show that \textbf{human brain speeds up the recognition of repeatedly presented objects due to faster memory encoding and accessing procedures on activated neurons}.
The cognitive neuroscience research~\cite{gazzaniga2006cognitive} also reveals that the priming effect is related to the long- and short-term memory of human brains: recent and just seen objects are stored in fast memory (short memory) and faster to be recognized than infrequently seen objects.
In a nutshell, human brains seem to be born with a kind of \textit{semantic} cache mechanism.

\begin{figure}[t]
\centering
\subfloat[][The priming effect]{\includegraphics[width=0.9\linewidth]{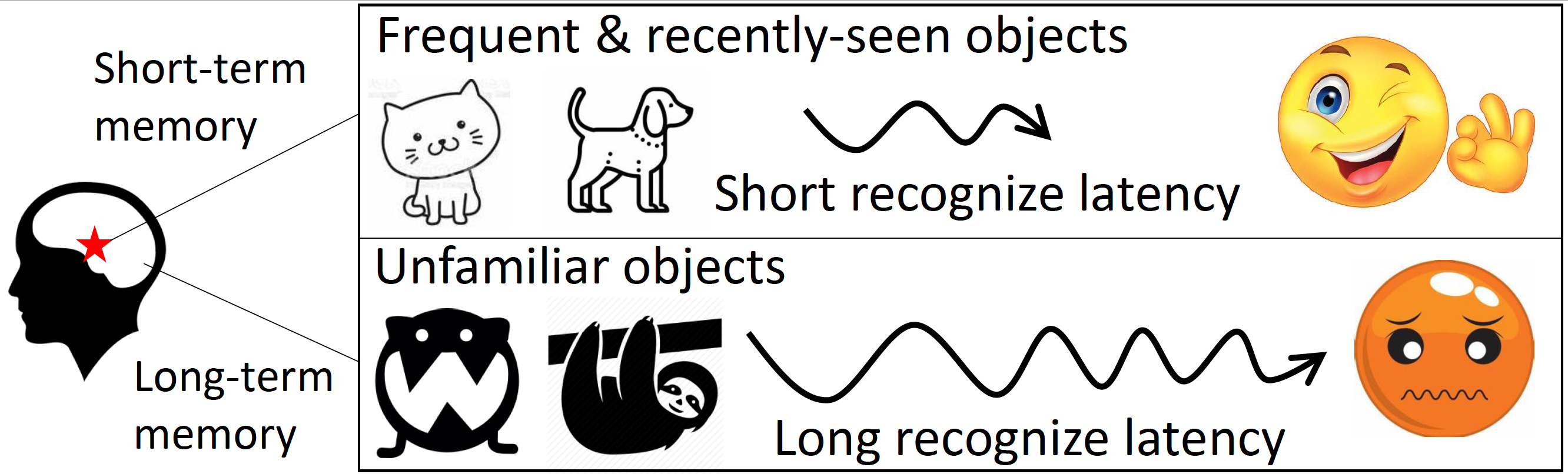}\label{fig:oper_collect_1}}\\
\subfloat[][The proposed method]{
\includegraphics[width=1\linewidth]{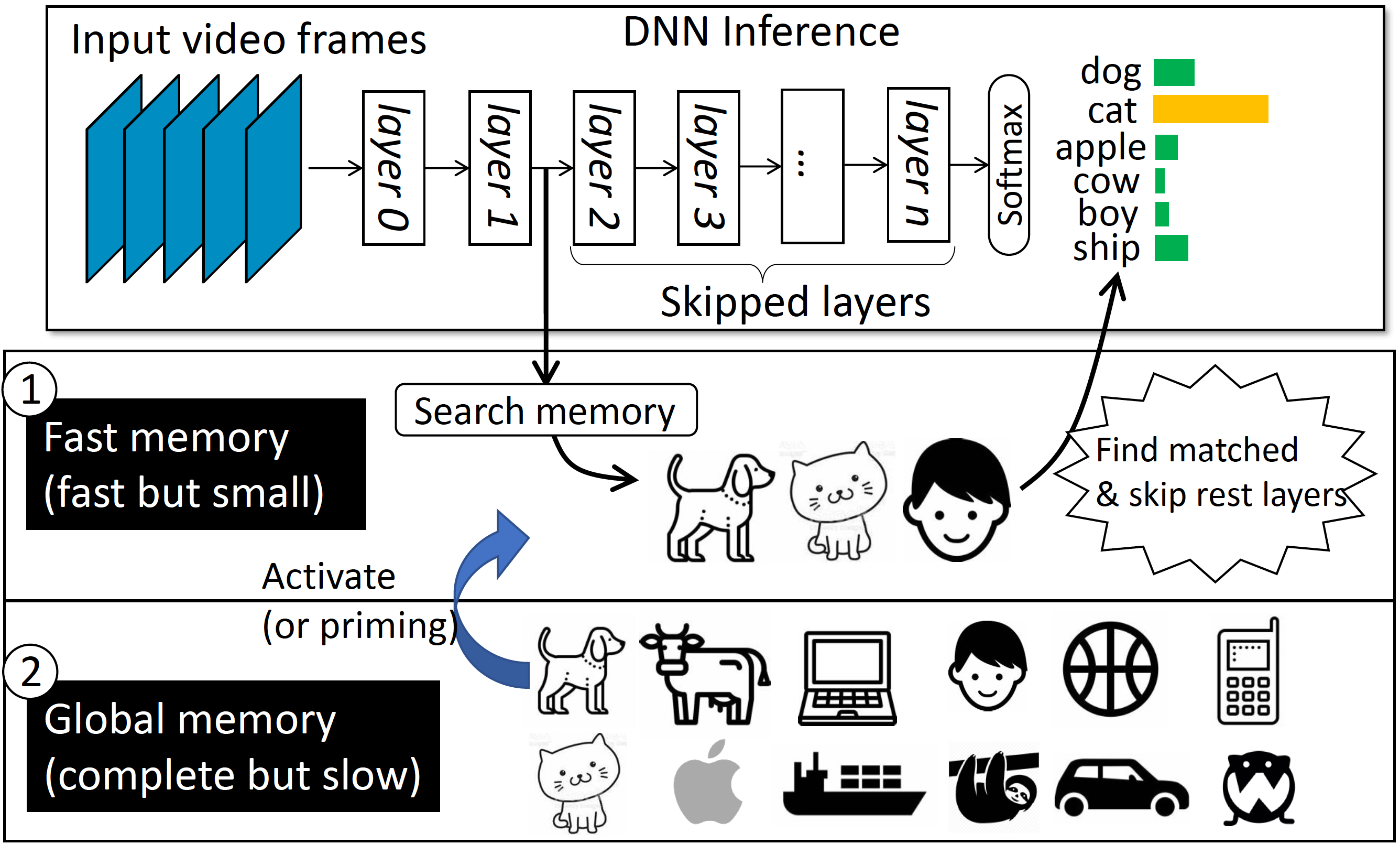}\label{fig:oper_collect_2}}
\caption{Overview of the proposed semantic memory (b) which inherits its key spirits from priming effect, a psychology phenomenon in human brains (a).
}
\label{fig:overview}
\end{figure}

A few recent studies have tried to exploit the opportunity of temporal redundancy.
However, unlike human brains that focus on semantic, high-level visual information, those systems only consider low-level visual information (either image pixels or blocks) by matching the input images~\cite{huynh2017deepmon} or intermediate feature maps (activations)~\cite{xu2018deepcache}.
As a result, the memory efficiency can be easily compromised by scene variation, e.g., object movement or light condition.
Moreover, caching images or feature maps incurs high computation and memory overhead due to their high dimensions.

\textbf{Our proposal}
We propose \underline{s}e\underline{m}an\underline{t}ic \underline{m}emory (\sys), a novel memory mechanism to accelerate CNN-powered mobile vision by infusing the priming effect with CNN inference.
Figure~\ref{fig:overview} shows the key idea of \sys with a hierarchical memory architecture. \sys promotes the most frequently- and recently-seen objects in the fast memory. During the CNN inference process, \sys looks up the fast memory per layer. Mimicking the priming effect, once the fast memory has a matched object with similar features, \sys skip the rest layers and directly output the prediction result.

However, orchestrating semantic cache with CNN inference is non-trivial and faces three major challenges.
(1) Directly looking up images or feature maps is cumbersome, e.g., taking about 10ms even with GPU acceleration~\cite{xu2018deepcache}, which can easily devour the benefits of our system.
Thus, an accurate yet low-cost memory encoding is desired to reduce the data dimension.
(2) The traditional execution flow of CNN inference can not directly obtain speedup by reusing semantics. An acceleration method is desired to leverage the hot-spot memory, which shall save computation without compromising model accuracy. 
(3) Temporal redundancy comes with dynamics: the characteristics of input images (feature distribution, object classes, etc.) may change over time due to the movement of user/camera.
Such dynamics can inevitably invalidate the memory results and finally compromise its performance. 

To address the first issue, we propose \textit{semantic vectors}, a memory encoding method that extracts the high-level vision information from intermediate feature maps during inference.
We present detailed analysis and demonstrate two preferred characteristics of semantic vector:
1) it is much more light-weight than directly caching and looking up feature maps, and
2) it is an effective indicator to accurately differentiate different object classes.

To address the second issue, we propose an `early exit' method that skips CNN layer's execution by exploring the temporal redundancy through matching semantic vectors. Determining the timing to exit the inference plays a critical role in making a trade-off between latency and accuracy. Based on our observation of the feature characteristics in CNN, we propose a cross-layer cumulative similarity to measure the confidence of early exit. 

To address the third issue, we introduce an adaptive and hierarchical priming memory, where we cache the hot-spot objects in the fast memory while leaving the complete set in the global memory. As the data distribution of the scene is not known in advance, we propose a cache replacement policy that takes frequency and recency of the observed data to predict the recurrence probability of each class in the future. Moreover, we propose an adaptive cache size and an adaptive semantic center to increase cache hit ratio and recognition accuracy under various and high-moving scenarios.  

We prototyped \sys on commodity CNN inference engine and comprehensively evaluated its performance on 5 popular CNN architectures, 2 large-scale video datasets, and both mobile CPU/GPU hardware.
The results show that \sys achieves 1.2--2.0$\times$ speedup and 13.7\%--48.5\% energy saving over no-memory method, and 1.1-1.5$\times$ speedup over prior cache systems~\cite{huynh2017deepmon,xu2018deepcache}.
Doing so, \sys incurs very low accuracy loss (1.05\% on average) and memory footprint overhead (2MB on average).
With the proposed subconscious recognition, our method even achieves higher accuracy than baselines.

We summarize our major contributions.
\begin{itemize}
    \item Semantic memory, a novel cache mechanism borrowed from neuroscience research, to accelerate on-device CNN inference.
    \item Three concrete techniques to take the semantic memory into effects: an accurate yet low-cost memory encoder, an early exit method, and an adaptive priming memory policy.
    \item A prototype of \sys on commodity CNN engine and extensive experiments showing its effectiveness.
\end{itemize}

\section{Inspiration and Challenges}

In this section, we present the inspiration for our ideas and challenges to cache semantic for mobile vision.

\subsection{Lessons from cognitive neuroscience}
\label{sec:CN}

Convolutional neural network is initially designed to imitate optic neurons of mammals due to high recognition accuracy on processing computer vision data \cite{lecun2015deep}. However, today's CNN models are of high computation complexity and energy intensity, putting a lot pressure on resource-constrained mobile and wearable devices. The human brain is much more efficient, which achieves high intelligence within a small power envelope.

The `priming effect' is a fundamental cognitive phenomenon and is born with the implicit memory in human brains \cite{gazzaniga2006cognitive}.
It refers to the changes in processing speed, bias or accuracy of one stimulus, following the same or related stimulus that has been recognized previously~\cite{primenature}.
For example, due to a prior experience of witnessing a cat, human's recognition of a cat becomes faster unconsciously in a short period.
The priming effect widely exists in lexical and vision cognitive processes \cite{primed, henson2003neuroimaging} and has been proved to be related to different anatomical regions in the brain. 

The priming effect represents a way of human brain taking advantage of temporal redundancy in continuous vision tasks. In the contrast, the typical way of CNN models doing inference is much less efficient because every input has to go through a full pass before making the final recognition, regardless of whether it has been witnessed frequently or recently before. Our work aims to accelerate CNN inference on mobile devices by introducing a fundamental mechanism that reduces computation workload on frequently- and recently-seen objects.

\subsection{Challenges}
\label{sec:challenges}

Although the objective of caching and reusing semantics to accelerate CNN execution is intuitive, we notice there are several major obstacles to build an effective memory mechanism like human brains.

\noindent $\bullet$\textbf{Efficient memory encoding against CNN models' over-parameterization (Section~\ref{sec:SC}).}
The underlying representational power of today's CNN models comes from the huge parameter space which results in an extremely large volume of intermediate data, i.e., feature maps, generated in hidden layers. Processing these feature maps requires a large memory footprint and high computation cost, which impedes fast memory encoding and accessing. 
An accurate yet low-cost memory encoding is desired to represent the high-level vision semantics from CNN layers' feature maps.
\sys introduces GAP function as the encoding tool to generate memory encoding from multi-dimensional feature maps.

\noindent $\bullet$\textbf{Obtaining speedup by high-level vision semantics (Section~\ref{sec:EE}).}
Previous memory designs~\cite{xu2018deepcache, huynh2017deepmon} mainly cache the hot-spot by using \textit{low-level} vision information, e.g., measuring pixel-level similarities.
However, human brain makes recognition of an object by its \textit{high-level} features instead of pixels' digital values.
Accelerating CNN inference by leveraging the high-level semantics requires a co-design of the proposed memory encoding and CNN's execution flow.
\sys demonstrates the feasibility of exiting inference early on intermediate CNN layers by using high-level semantics.

\noindent $\bullet$ \textbf{Battling dynamics on scenario variation (Section~\ref{sec:EEM}).}
On mobile or wearable devices, the scene may change drastically from time to time with the movement of the user/camera.
Such a high dynamics raises two issues.
First, the data distribution and the scene complexity is not known in advance.
For example, an auto-driving car mainly needs to recognize cars on highways but also has to deal with a much larger number of object classes on downtown streets, e.g., traffic lights, pedestrians, stop signs, etc.
Second, the characteristics of real scenario data may differ from the training set on which the model is trained.
This data variation may result in accuracy loss with more real-scenario data accumulated.
To tackle the above challenges, we propose two techniques, 1) an adaptive cache size to adjust for different scenarios, and 2) an online method to update semantic vectors with the input image.

\section{System Design Overview}

\begin{figure}[t]
	\begin{center}
		\includegraphics[width=1\linewidth]{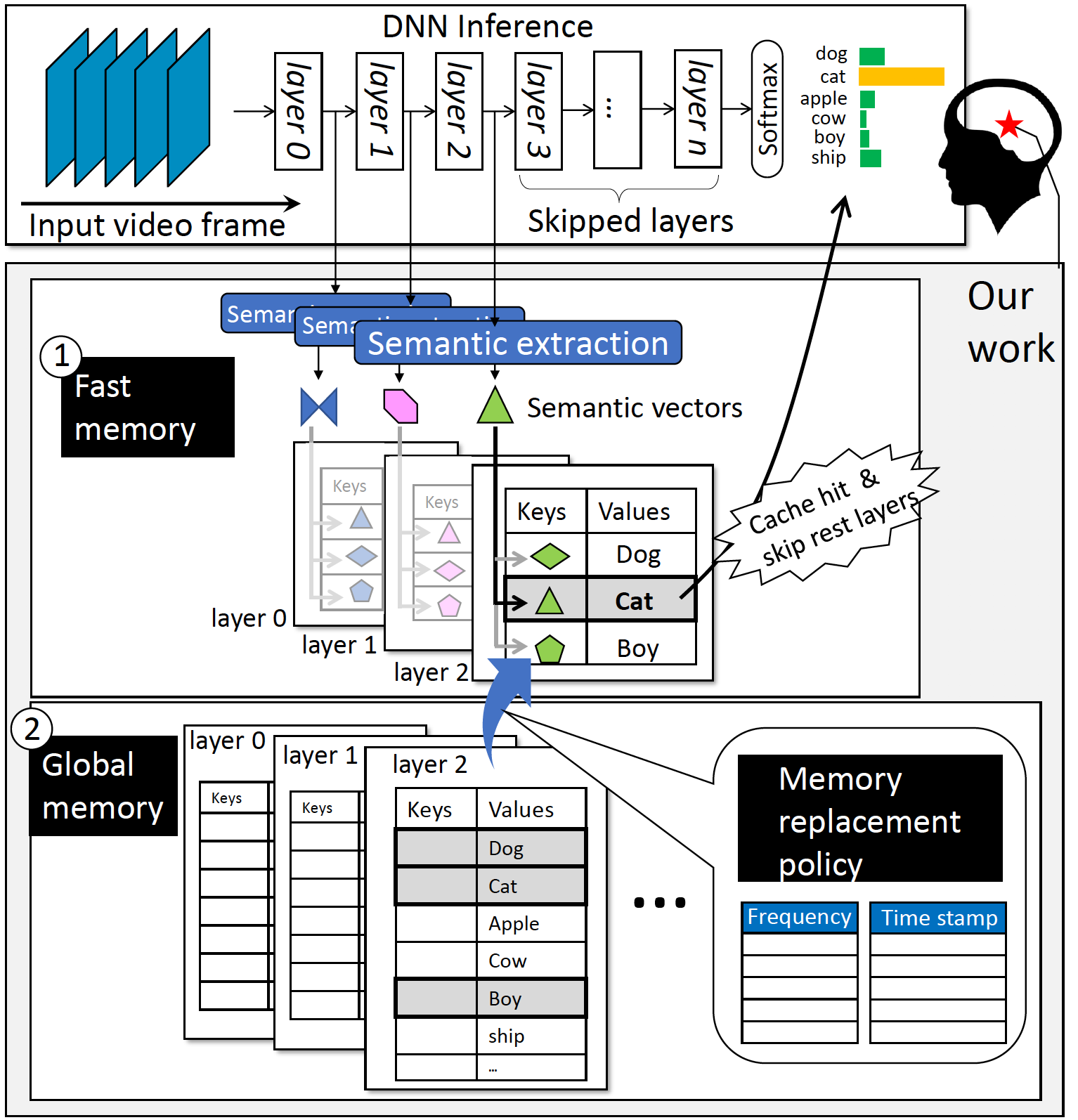}
	\end{center}
	\caption{The workflow of \sys. 
	}
	\label{fig:pipeline}
\end{figure}

\sys speeds up CNN inference by skipping some layers' execution according to the cached (activated) memory of frequently and recently-seen objects.
The key advantages of \sys include: 1)\textit{Small memory footprint and low memory lookup cost.} Instead of directly storing the multi-dimensional feature maps, \sys only uses a set of vectors as memory encoding to encode high-level vision information for each category. The memory encoding has a reduced dimension and thus takes much less memory and memory lookup cost.  
2)\textit{Enjoy pervasive AI hardware acceleration.} \sys is capable to directly skip CNN layers. In contrast to pixel/region reuse~\cite{huynh2017deepmon, xu2018deepcache}, \sys does not require any modification to the original convolution operator, and can directly use existing mobile AI hardware for acceleration, e.g SIMD units or GPUs. 3)\textit{Soft constraints on the matching object.} Different from low-level vision (pixel or region) reuse, \sys explores temporal redundancy according to high-level vision information. The reused objects only need to be of the same category that has high-level feature similarity, but not have to follow pixel-level similarity constraints. This helps to discover more temporal redundancy and enables more acceleration chances. 4)\textit{Robust on scene variations.} \sys propose an adaptive policy to adjust the priming memory mechanism for different types of variations in a mobile device.

\textbf{Workflow.} Figure \ref{fig:pipeline} shows the overall workflow of \sys. 
\sys introduces a global memory and a fast memory to improve the process of traditional CNN inference. First, \sys employs a \textit{global memory} to cache the frequency and timestamp of all classes, as well as the feature expression of each class extracted in the training set.
Second, \sys uses a \textit{fast memory} to cache a few hot-spot classes and their features for fast matching. The caching replacement policy predicts the possibility of objects' occurrence in the mobile video streams according to two observations: 1) a long-tail distribution: some objects are much more frequently seen than others. 2) temporal locality: a recently-seen object is more likely to appear in the next few frames.
During the CNN inference process, \sys extracts the intermediate feature per layer and matches them with the cached features in fast memory. Once matched, \sys skips the rest of the layers and directly outputs the final results. At last, \sys updates the frequency table and time-stamp table, which will be used in the memory replacement policy periodically.

\sys proposes the following technologies to solve the challenges mentioned before. 

\textbf{Semantic memory encoding: perform memory encoding and lookup. (Section \ref{sec:SC})} \sys encodes the large volume of feature maps to low-dimensional \textit{semantic vectors} in memory. It does so based on the global average pooling (GAP) \cite{lin2013network}, a widely used function in person Re-ID tasks \cite{wei2017glad, yao2019deep} as a dimension reduction and high-level feature extraction operation to perform person matching. We present a detailed analysis of semantic vectors' grouping performance that is used to distinguish different classes' categories. 

\textbf{Early exit: obtaining speedup with a novel metric. (Section \ref{sec:EE})} As CNN performs inference layer by layer, \sys calculates the semantic vector on each layer's output feature map. Then, the semantic vector works as the `keys' to lookup corresponding objects' categories as `values' by measuring \textit{cosine similarities}. To increase the robustness of prediction accuracy, \sys proposes the \textit{cross-layer cumulative similarity} to determine the exiting timing by combining the confidences of multiple layers.

\textbf{Adaptive Priming Memory: cache and update the semantics of mobile video frames. (Section \ref{sec:EEM})} 
To reuse semantics efficiently to deal with the scene variation in mobile video, \sys sets a memory replacement policy in the global memory, which maintains a frequency table to record the time of each object presented in the history, and a time-stamp table to record the consecutive non-appearing frames of each object recently. Then, the cache replacement policy will calculate a score based on the current frequency table and the time-stamp table to select the object classes with the highest recurrence probability, and cache them in the fast memory. To overcome the high dynamic scenarios in mobile vision video, \sys introduces the adaptive cache size and the adaptive semantic centers. A probability estimation method is adopted to tune the cache size in fast memory based on the current frequency table and time-stamp table. To make the semantic centers constantly adapt to test scenarios, \sys gradually update the semantic centers by accumulating the semantic vector in a weighted average manner.

\section{Semantic Memory Encoding}
\label{sec:SC}

An efficient memory encoding is a prerequisite because the large volume of feature maps introduces high computation cost as well as large memory footprint overhead.
It finally impedes fast memory encoding and accessing for a priming effect.

Besides being fast, the memory encoding must accurately capture the key features of the corresponding objects so that the image semantics can differentiate different classes.
We use \textit{separability} to evaluate the memory encoding on classifying different memories.
During inference runtime, we adopt the metric learning method \textit{cosine similarity} \cite{nguyen2010cosine} to measure the separability between memories and the semantics of the new input layer by layer. 

Next, we first introduce \textit{semantic vector} that we use to encode memory. Then, we evaluate the separability within semantic vectors of each layer. 

\subsection{Semantic vectors}

\begin{figure}[t]  
    \setlength{\belowcaptionskip}{-0.5cm}
	\centering
	\includegraphics[width=1\linewidth]{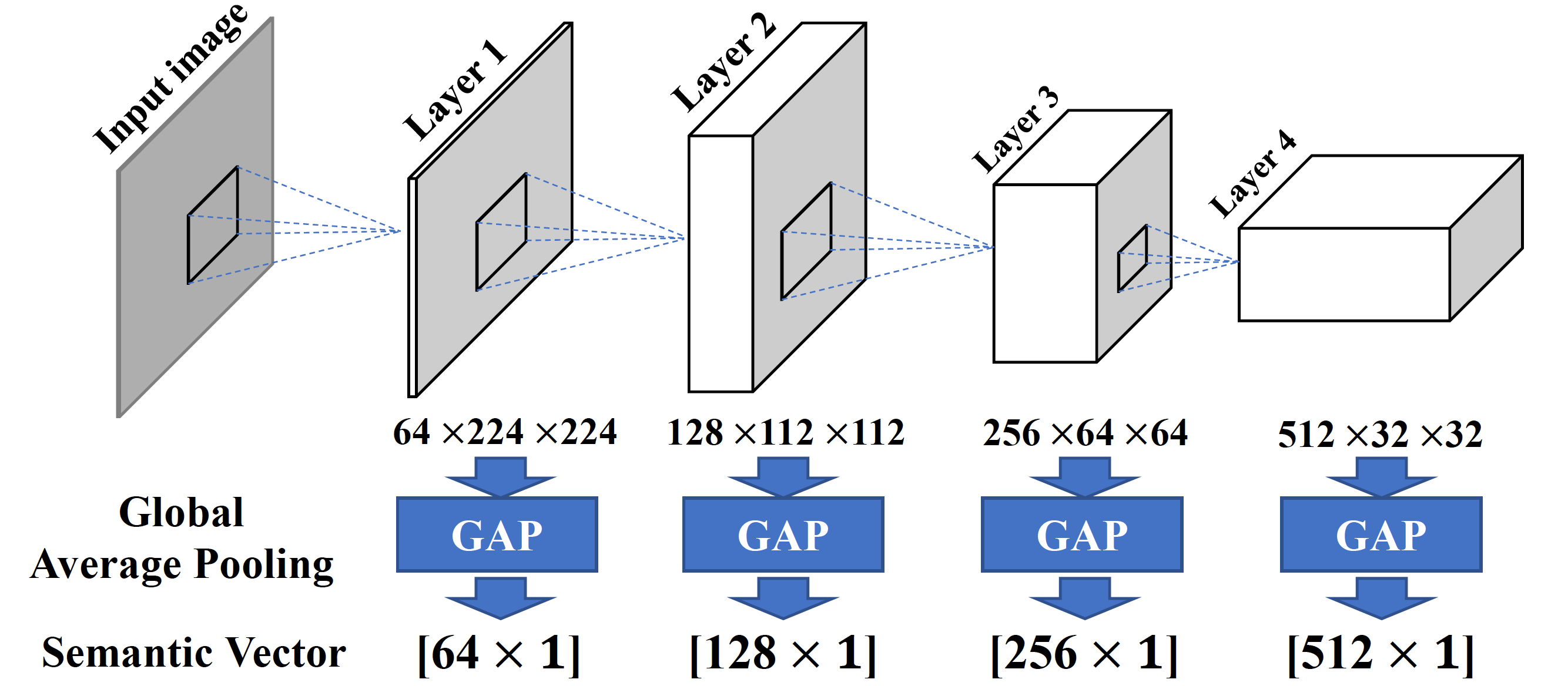}
	\centering
	\caption{Semantic vectors extraction.}
	\label{GAP}
\end{figure}

\sys memorizes intermediate data during CNN inference. An early exit happens on the condition that some layer's feature map matches with the memory of some previous inputs. To efficiently memorize the intermediate data across CNN layers, we propose \textit{semantic vector} that is retrieved by applying a global average pooling (GAP) function~\cite{lin2013network} on feature maps. The global average pooling takes the average of each feature map and outputs a result vector. As shown in Figure \ref{GAP}, the GAP has applied to each layer' (modules')s output feature maps. 

Widely used in person Re-ID tasks~\cite{wei2017glad, yao2019deep}, global average pooling (GAP) serves as a dimension reduction and key feature extraction to perform person matching by measuring vector distances. Similar to person Re-ID tasks, we use semantic vectors as IDs (or keys) for object classes. By measuring the similarity between semantic vectors, we can establish the mapping between each individual's semantic vector and object classes. 

Semantic vectors have preferred characteristics of small memory foot-print and low computational cost because it greatly reduces raw feature maps' dimensions. Given a feature map size of $C\times H\times W$, the semantic vector is only $C$, where $\langle C, H, W\rangle$ stands for channels, row, and column, respectively. Despite that, the semantic vectors also have good separability, which makes it easier to differentiate objects of various classes from semantic vectors. We provide a detailed analysis in the following subsection.

\subsection{Rationales}
\label{sec:separability}

\begin{figure*}[t!]
\centering
\subfloat[][layer 4]{\includegraphics[width=0.3\linewidth]{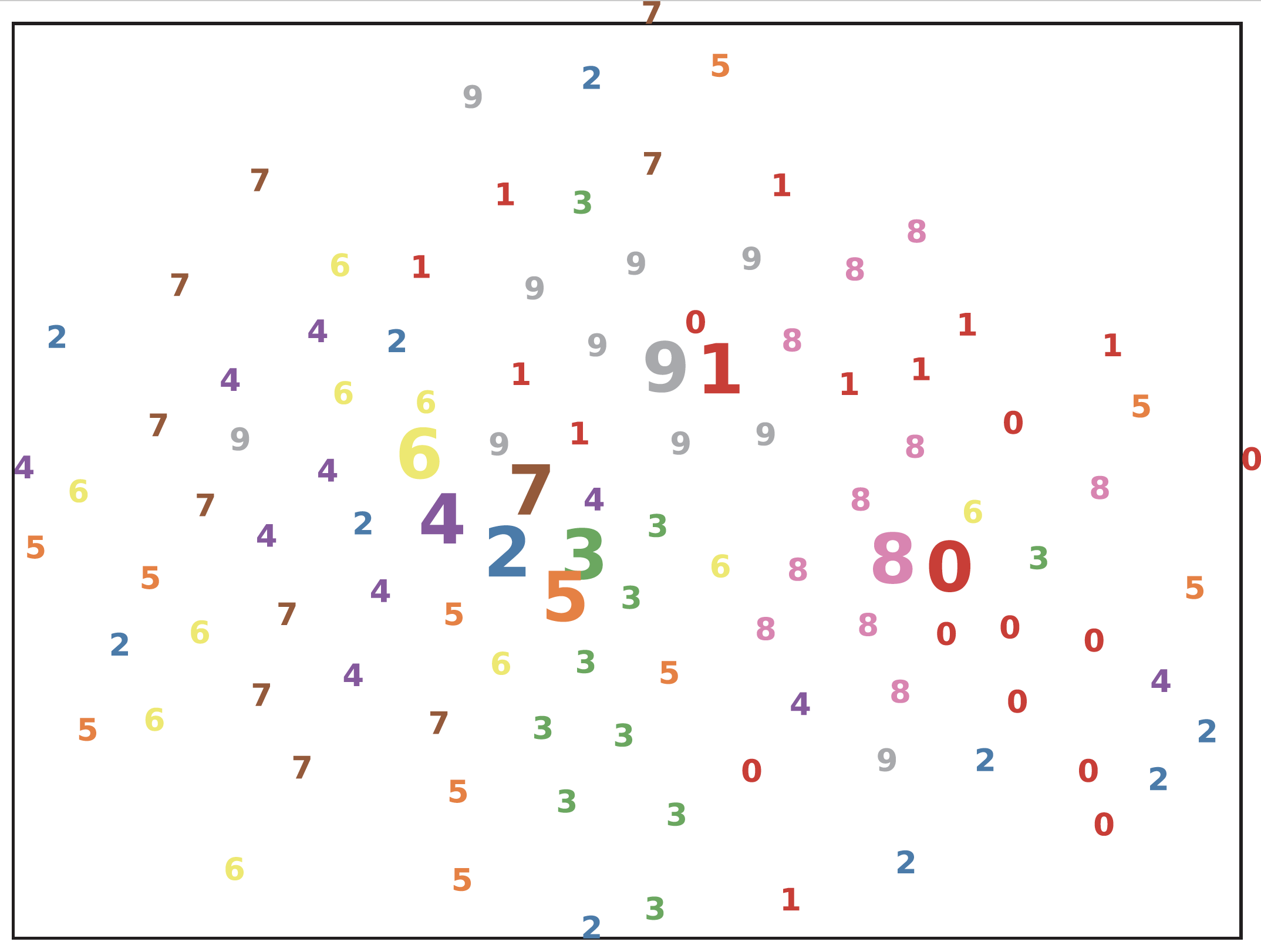}\label{fig:sep_1}}
\subfloat[][layer 8]{
\includegraphics[width=0.3\linewidth]{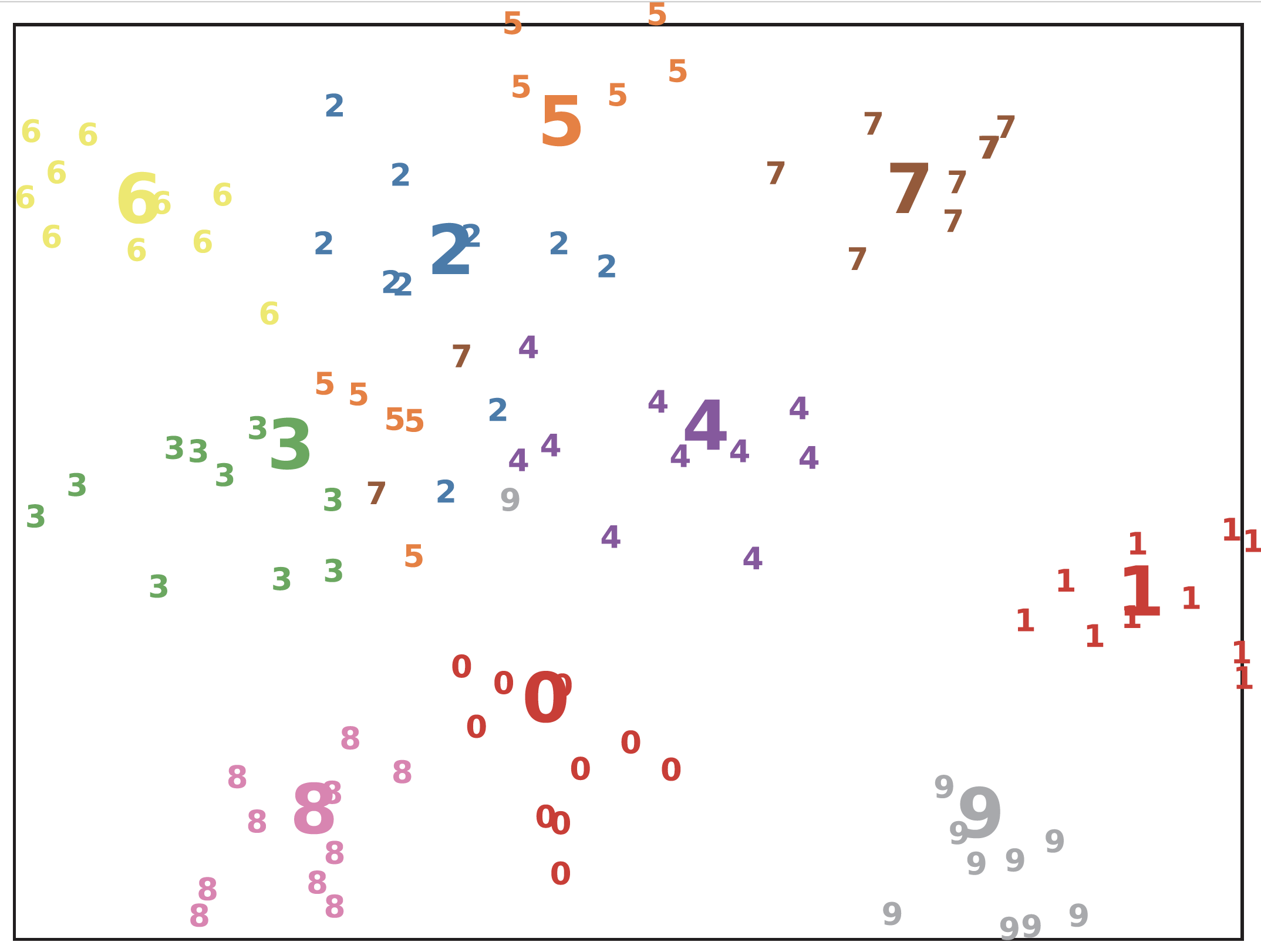}\label{fig:sep_2}
}
\subfloat[][layer 12]{
\includegraphics[width=0.3\linewidth]{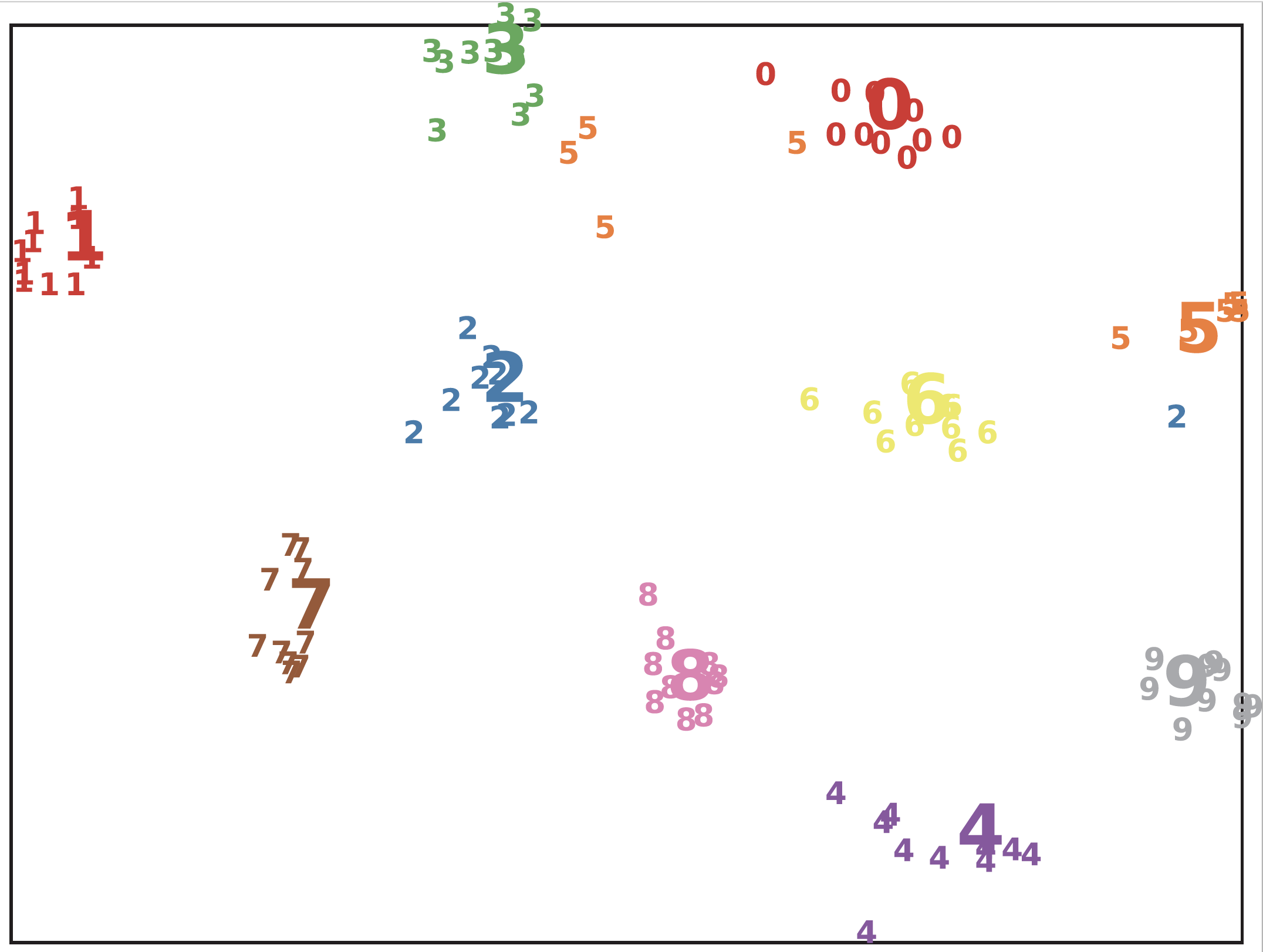}\label{fig:sep_3}
}
\caption{Visualized separability of semantic vectors for different VGG16 layers, showing that going deeper the semantic vectors can be more accurately separated.}
\label{fig:separability}
\end{figure*}

Intra-class distance and inter-class distance \cite{schroff2015facenet} are two key indicators to measure the semantic vectors' performance on clustering data. When the intra-class distance is smaller than the inter-class distance, we can distinguish different targets well. To analyze its separability of different objects, we sample three hidden layers' semantic vectors in the VGG-16 model and visualize them in Figure~\ref{fig:separability}. Since semantic vectors are still multi-dimensional, we use t-SNE method \cite{maaten2008visualizing} for visualization. In each subfigure of Figure \ref{fig:separability}, the bigger labels mean the semantic centers of each object class, and the smaller labels denote the semantic vector of test samples. We can draw two conclusions from Figure~\ref{fig:separability}. First, semantic vectors present clear separability in some hidden layers, such as layer 8 and layer 12 in Figure~\ref{fig:sep_2} and \ref{fig:sep_3}. Second, the separability becomes more obvious as the layers go deep, which leads to higher classification accuracy, by comparing Figure~\ref{fig:sep_1} to later layers. The observation above provides us an opportunity to distinguish different objects in the early layers of CNNs by measuring the distance between semantic vectors of new inputs and semantic centers of different objects in memory.

Based on the analysis above, we adopt the cosine distance \cite{nguyen2010cosine} as the metric to evaluate the distance of different objects.  
For a new frame, we first encode its semantic vectors layer by layer.
Let ${SV}^{l}$ ($l\in \left [ 1,L \right ]$) denotes the extracted semantic vector at layer $l$, in which $L$ is the number of convolutional layers or building blocks (such as bottleneck \cite{he2016deep} and Inception \cite{szegedy2015going}). 
Let ${SC_{j}}^{l}$ ($j\in \left [ 1,n \right ]$, $l\in \left [ 1,L \right ]$) denotes the semantic centers of object $j$ at layer $l$. 
Then, we measure the separability of semantic vectors in a single layer. We calculate the cosine similarity between the extracted semantic vector ${SV}^{l}$ and the semantic centers of objects in the memory. The similarity in layer $l$ can be formulate as:
\begin{equation} 
s_{j}^{l}=\xi\left ( SV^{l},SC_{j}^{l} \right )\in \left [ -1,1 \right ],j \in \left [ 1,n \right ],
\label{eq:cosine}
\end{equation} 
in which, $\xi \left ( \cdot  \right )$ is the cosine similarity function, $s_{j}^{l}$ is the similarity between the semantic vector of new input and the semantic center of object $j$ at layer $l$. The larger value means higher similarity.
Based on the relationship between intra-class distance and inter-class distance, if the similarity between the input and semantic centers of one object in the memory is significantly larger than the other objects, the separability is high. Therefore, the separability in a single layer $l$ can be measured by:
\begin{equation} 
sep^{l}=\frac{s_{H}^{l}-s_{SH}^{l}}{s_{SH}^{l}}.
\end{equation}
in which $s_{H}^{l}$ is the highest similarity result at layer $l$, which represents the object with the highest confidence, and $s_{SH}^{l}$ is the second-highest result. The larger the separability $sep^{l}$, the higher our confidence to make a distinction.
The separability will be applied and improved in our memory lookup period in the following section.

\section{Early exit}
\label{sec:EE}

\sys uses the semantic vectors to capture the repeatedly seen objects and save computation workload by the early exit. 
During the CNN inference, we first encode the semantic vectors layer by layer and then match them with the semantic centers of objects in memory. If there is enough confidence about the matching results, the CNN inference will exit early and the rest layers are skipped directly. The semantic centers are initialized with a grouping center (the average of semantic vectors) on the training dataset and will be updated during runtime, which will be discussed later.

Based on the observation from Figure \ref{fig:separability}, the semantic vector's separability in shallow layers is not as strong or stable as the deeper layers. To exit the inference as early as possible while ensuring inference accuracy, we adopt the cross-layer cumulative similarity to evaluate memory matching results during the memory lookup period. Overall, the memory lookup can be divided into three steps. 

\textbf{Step 1.} For the current input frame (image), we adopt global average pooling to encode the memory and generate semantic vectors layer by layer during the CNN inference. 

\textbf{Step 2.} At each layer, we leverage cosine similarity \cite{nguyen2010cosine} to lookup the most matched semantic centers in memory. The matching level is represented by similarity $s_{j}^{l}$ in Eq.\ref{eq:cosine}. To improve the robustness to distinguish all objects in a single layer, we introduce the cross-layer cumulative similarity to evaluate memory matching result:
\begin{equation} 
SA_{j}^{l}=\sum_{l_{0}=1}^{l}s_{j}^{l_{0}}\times weight_{l_{0}}, j\in \left [ 1,n \right ],
\end{equation}
where $SA_{j}^{l}$ is the similarity accumulation result between the semantic vector of the current frame and semantic center of the object $j$ in memory from layer 1 to layer $l$, $weight_{l_{0}}$ is the weight of the results in the $l_{0}$-th layer. 
Considering the separability will become more and more obvious as the CNN layer going deep, the preceding equation requires a sequence of increasing weight values. To this end, we adopt a \textit{exponential} ($weight_{l_{0}}=2^{l_{0}-1}$) weighted decay. As this exponential function has a useful characteristic ($\sum_{l_{0}=1}^{l-1}2^{l_{0}-1}=2^{l-1}-1=weight_{l}-1$), making the weight of current layer $l$ and the cumulative weights of previous shallow layers almost equal. This weighting method not only ensures that deeper layers are assigned greater weights but also fully considers the similarity results on the entire inference path.

\begin{figure}[t]
	\centering
	\includegraphics[width=1.0\linewidth]{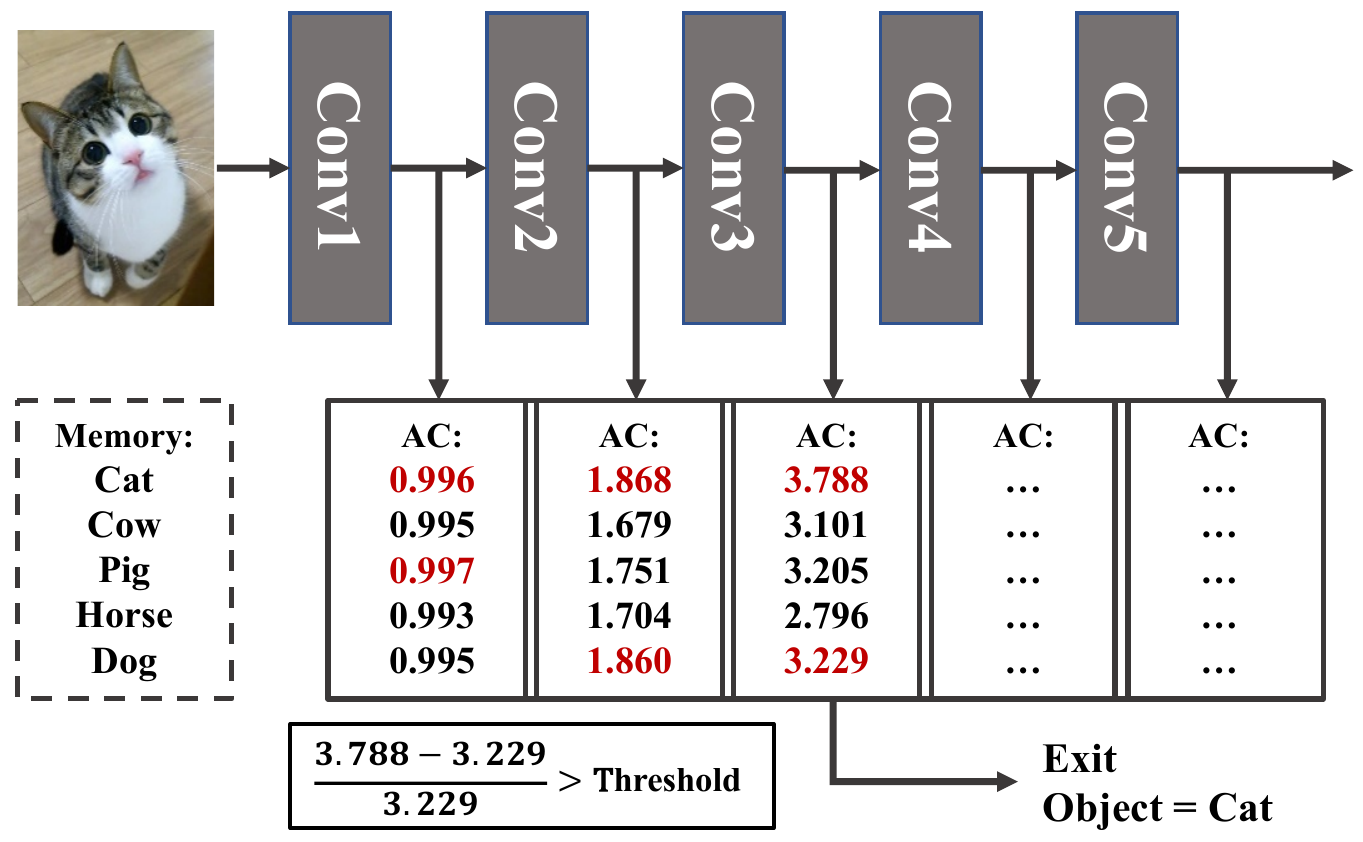}
	\centering
	\caption{Memory look up by accumulated confidence (AC) metric. Memory: objects in memory.}
	\label{confidence}
\end{figure}

\textbf{Step 3.} Based on our observation and analysis in Section \ref{sec:separability}, 
we measure the separability across layers by the distance between the highest similarity accumulation result and other similarity accumulation results in the current layer. 
Let $SA_{H}^{l}$ be the highest similarity accumulation result in layer $l$, which represents the object with the highest matching level. $SA_{SH}^{l}$ be the second-highest result in layer $l$, which is the biggest interference term to the memory matching. Then, we measure the accumulated confidence (AC) score of early exit (separability across layers) in the $l$-th layer as follows:
\begin{equation} 
AC^{l}=\frac{SA_{H}^{l}-SA_{SH}^{l}}{SA_{SH}^{l}}.
\end{equation}
The larger the above score, the higher the confidence we have about the memory matching. An example is shown in Figure \ref{confidence}. 
A global threshold $\tau $ is set, and if the score $AC^{l}$ exceeds the threshold $\tau $, \sys will exit the CNN inference at layer $l$ and output the corresponding matching result of $SA_{H}^{l}$.

\section{Adaptive Priming Memory}
\label{sec:EEM}

\sys uses a hierarchical memory organization, which is composed of fast memory and global memory.
The global memory contains a complete set of semantic centers of all objects, while the fast memory caches a subset of the frequently- and recently-seen objects.
During inference, \sys only looks up the fast memory. 

We propose three techniques to overcome the highly dynamic scenarios in mobile vision tasks. A frequency table and a time-tamp table are maintained in the memory replacement policy, which is used to update the fast memory periodically. For the scene dynamics, we propose to tune the fast memory size adaptively according to the scenarios. For the semantic dynamics, we propose to adjust the semantic center incrementally after every inference.

\subsection{Cache replacement policy}
\label{sec:CRP}

The goal of our cache replacement policy is to update the fast memory for efficient semantic lookup. Members in fast memory are replaced according to a policy combining the frequency and recency of the objects presented in the video stream. Thereby, \sys maintains a \textit{frequency table} and a \textit{time-stamp table} for cache replacement.

\textbf{Frequency table} keeps a record of the number of times that each object class presented in history. It is initialized as zeros and updated by every inference output during runtime. A class with a high score in the frequency table means it has a high probability of witnessing throughout the entire `history', such as `cars' for downtown street cameras. Thus, the corresponding class should be given higher priority when promoting it from global memory to fast memory.

\textbf{Time-stamp table} keeps a record of the recency of each object class. The intuition is that the most recently seen objects are likely to appear again in the next few frames. The time-stamp table works like a forgetting mechanism~\cite{ebbinghaus2013memory}, where all object classes decay with time. Upon every inference complete, all other classes that have not been witnessed on a time interval will be decayed by a certain ratio so that most recently-seen have the highest score at present. In our experiment, the forgetting mechanism is defined as $\psi _{i}=\psi _{i}\times \left ( 0.25 \right )^{\left \lfloor \frac{TS_{i}}{W} \right \rfloor}$, in which $W$ is the the size of observation time window. According to the values in the time-stamp table, we decay the memory (the effect of the corresponding frequency on the cache updating) every consecutive $W$ frames.

\textbf{The replacement policy} takes the Top-k highest score that are calculated by the following equation to select the objects from the global memory and cache them in the fast memory,
\begin{equation}
    Score_{i}=FT_{i}\cdot\left ( 0.25 \right )^{\left \lfloor \frac{TS_{i}}{W} \right \rfloor}, i\in \left [ 1,n \right ],
\end{equation}
where $FT_{i}$ is the frequency of object $i$ in the frequency table, $TS_{i}$ is the consecutive non-appearing frames of object $i$ in the time-stamp table.

This equation lets fast memory cache the constantly-often-seen and the most-recently-seen objects. For the frequently-seen but NOT recently-seen objects, its overall score will be degraded due to a high decay ratio by the time-stamp table, and vice versa. For objects that are NOT frequently-seen and NOT recently-seen, it has the least possibility to be cached. With this policy, \sys can enjoy inference speedup by keeping the hottest classes in fast memory.

\subsection{Adaptive cache size}

Due to the mobile scene's drastic data variation, the number of hot-spot classes may vary a lot under different scenes. 
Using a large fast memory (cache) for a simple scenario causes overhead on memory retrieving. Thus, \sys adopts a probability estimation method to figure out the optimal cache size.

Considering that the frequency table and the time-stamp table can reflect the frequency and the recency of each object, we use the two tables as the observed data to estimate the reappear probability of all the objects in the memory in the future. Specifically, we adopt the percentage of each object's score (calculated in the cache replacement policy) in all objects to estimate their reappearance probability. Let $\Phi $ denotes the set of all objects in the global memory and $\Psi $ be a subset of $\Phi $ which are selected based on the Top-k highest scores in cache replacement policy. Suppose the next video frame belong to object $\theta$, then the probability of event $A=\left \{ \theta \in \Psi \right \}$ can be formulated as:
\begin{equation}
P\left ( A \right )=\sum_{i=1}^{k}\frac{Score_{i}}{\sum_{i=1}^{n}Score_{i}}, 
\label{eq:adaptive}
\end{equation}
in which $k$ is the number of cached objects in fast memory, $n$ is the number of objects in the global memory. According to the statistics, if $P\left ( A \right )$ can exceed the most commonly used confidence level (CL) 95\% \cite{dekking2005modern, zar1999biostatistical}, we can believe that the event will happen with a high probability. Therefore, based on Eq. \eqref{eq:adaptive}, we can adaptively tune the cache size of fast memory after each inference to overhead the scene changes. The experiments in Section \ref{Eva:adaptive} have shown that such a technology can bring 21.6\% hit ratio improvement.

\subsection{Adaptive semantic centers}

The semantics center works as the key for memory lookup, which is by measuring semantic vectors' similarity to the semantic center. An improper semantic center may finally degrade memory lookup accuracy. However, training dataset's biases against the data in the real world may generate an improper semantic center. The scene variation may also cause a drifting optimal semantic center from time to time. To this end, we propose an adaptive semantic center method so that it can be continuously updated according to the semantics extracted from the real-world data.

\sys initially warms up the semantic centers using the training data. During runtime, we gradually update the semantic centers by accumulating the semantic vector in a weighted average manner. For the current frame, if the predicted result is object $j$ and the CNN inference stop at layer $l$, then the semantic centers of object $j$ before layer $l$ will be updated as follow:
\begin{equation}
{SC_{l_{0}}^{j}}'=\frac{SC_{l_{0}}^{j}\cdot m_{l_{0}}^{j} + SV_{l_{0}}^{j}}{m_{l_{0}}^{j}+1}, l_{0}\in \left [ 1, l \right],
\end{equation}
in which, $SC_{l_{0}}^{j}$ denotes the original semantic center of object $j$ at layer $l_{0}$, ${SC_{l_{0}}^{j}}'$  denotes the new semantic center of object $j$, $SV_{l_{0}}$ is the encoded semantic vector of new example, $m_{l_{0}}^{j}$ is the update times of object $j$ at layer $l_{0}$, which includes the update times from the training dataset and the test scenario. 
By doing so, the semantic centers in global memory can be adjusted incrementally after every inference to overcome the semantic dynamics brought by highly dynamic scenarios in mobile vision tasks.

\section{Implementation}

We prototype \sys atop ncnn \cite{ncnn}, an open-source deep neural network inference computing framework optimized for mobile platforms.
The ncnn provides a set of APIs for easy interaction during the model forwarding, allowing extraction of intermediate layers with relatively low overhead.
While \sys currently supports mobile CPU and GPU, it can be easily extended to more device types, e.g., DSP and NN accelerators.
The implementation was split into two main parts: pre-processing of model files, and run-time inference. For the pre-processing step, we develop a tool that parses a converted ncnn model into graph presentation, inserts global average pooling (GAP) layers to predefined locations, and reconnects the graph based on requirements of ncnn by inserting split layers.
Overall, the implementation of \sys contains 4200 lines of C++ codes.

While \sys adopts the same cache/reuse strategy for different devices, e.g., CPU and GPU, we further tune the extraction module based on hardware characteristics to further reduce the cache overhead.
On CPU, we make a shallow copy of the tensor on the target extraction layer and forward through a global average pooling layer to get the feature vector.
On GPU, we implement a zero-copy data path based on Vulkan API \cite{Vulkan}, allowing tensor extracted from the network to be fed into our global averaging pooling layer without going through CPU memory.

Noting that, \sys is more instrumentation-friendly as compared to prior cache mechanisms~\cite{xu2018deepcache,huynh2017deepmon}, because those methods require to revise the neural layer implementation (kernels).
For example, ncnn has more than a hundred different implementations for convolution operation, with nearly one hundred thousand lines of code.
By contrast, \sys directly skips some complete operations in CNN and does not require any modification to the convolution calculation during inference, which can be easily applied to all the existing deep learning frameworks on mobile/wearable devices.

Our prototype is fully compatible with any existing ncnn models and applications, thus incurring zero overhead to developers. Besides, we expose key parameters, e.g., $\tau$, exposing rich accuracy-latency trade-off to developers so it can flexibly fit into task-specific requirements.
The relationship between threshold, accuracy, and latency will be given in Section \ref{parameter}.

\section{Evaluation}

In this section, we comprehensively evaluate \sys on diverse models, datasets, and metrics.
Overall, the results show that our method can outperform existing systems by a large margin.

\subsection{Experimental setup}

\textbf{Evaluation platform.}
We evaluate \sys on a Google Pixel 4XL mobile device,
which is equipped with a Qualcomm Snapdragon 855 Mobile SoC and 6GB LPDDR4x memory.
Snapdragon 855 is a big.LITTLE SoC consisting of four big Cortex-A76 cores, four little Cortes-A55 cores, and an Adreno 640 GPU.

\textbf{Benchmark Datasets.} We use two large-scale datasets UCF101 \cite{soomro2012ucf101} and long-tail CIFAR-100 \cite{krizhevsky2009learning} to evaluate the performance of \sys. 
\textbf{UCF101} is an action recognition dataset of realistic action videos, including 101 action categories. The dataset consists of 13,421 short videos. Following the settings of DeepCache \cite{xu2018deepcache}, we select 10 types as a subset for evaluation: \textit{Basketball}, \textit{ApplyEyeMakeup}, \textit{CleanAndjerk}, \textit{Billiards}, \textit{BandMarching}, \textit{ApplyLipstick}, \textit{CliffDiving}, \textit{BrushingTeeth}, \textit{BlowDryHair}, and \textit{BalanceBeam}. FFmpeg \cite{ffmpeg} is used to extract raw frames from those YouTube videos. Finally, 70,928 raw images are used. \textbf{CIFAR-100} is for object classification task, consisting of 60,000 images, and 100 object classes.

To evaluate the robustness of \sys against scene variation and its soft constraints on the reused objects, we adopt the long-tail CIFAR-100 as an extreme scenario. 
It is well-known that the frequency of object occurrence in natural scenes follows a long-tail distribution \cite{zhu2014capturing, salakhutdinov2011learning}.
Therefore, following the segmentation method in \cite {cao2019learning}, we split and shuffle the CIFAR-100 test dataset into 1,442 test images with long-tail distribution to simulate the object occurrence in natural scenes.
The resolution of the input images from UCF101 and CIFAR-100 are $224 \times 224$ and $32 \times 32$, respectively. In such rapidly changing scene, to avoid the effect of the high frequency value of some objects that have not appeared for a long time, we will decay the historical memory in the frequency table based on the forgetting mechanism in Section \ref{sec:CRP}.

\textbf{Models.} To verify that \sys is applicable to various types of CNN architecture, we use five widely-adopted network structures: AlexNet \cite{krizhevsky2012imagenet}, GoogleNet \cite{szegedy2015going}, ResNet50 \cite{he2016deep}, MobileNet V2 \cite{howard2018inverted} and VGG16 \cite{simonyan2014very}.
For action recognition, the first four models above are used, and VGG16 is adopted to the long-tail CIFAR-100.

\textbf{Evaluation metrics.} We use five metrics to comprehensively evaluate the performance of \sys: latency reduction (Section \ref{sec:latency}), accuracy loss (Section \ref{sec:accuracy}), memory overhead (Section \ref{sec:memory}), energy saving (Section \ref{sec:energy}), and early exit ratio (Section \ref{sec:exit}). 

\textbf{Alternatives.} We compare \sys with following alternatives: \textit{no-cache-CPU}, \textit{no-cache-GPU}, \textit{DeepMon} \cite{huynh2017deepmon} and \textit{DeepCache} \cite{xu2018deepcache}. \textit{no-cache-CPU/GPU} use mobile CPU/GPU to compute the complete CNN models without cache reuse. \textit{DeepMon} and \textit{DeepCache} are two state-of-the-art cache-based approaches. To make a fair comparison with DeepCache and DeepMon, we prototype \sys using the same inference engine (ncnn) with the same configuration as DeepCache~\cite{xu2018deepcache} without single instruction, multiple data (SIMD). For other experiments, if not otherwise specified, we prototype the framework on the ncnn with SIMD.

\subsection{Latency reduction} 
\label{sec:latency}

In this section, we evaluate the latency reduction on action recognition and classification when applying the proposed \sys. The latency is tested on different devices with configuration: mobile CPU without SIMD acceleration, mobile CPU with SIMD acceleration, and mobile GPU acceleration. 
To evaluate the performance on the entire dataset, we adopt \sys to accelerate the CNN inference for each input frame and calculate their average processing time.

\begin{figure}[t]  
	\centering
	\includegraphics[width=1\linewidth]{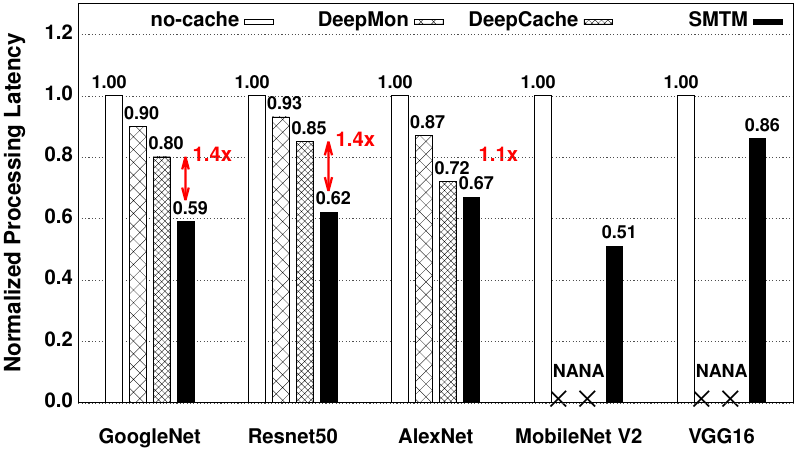}
	\centering
	\caption{Average processing latency with CPU (w/o SIMD) on action recognition (AlexNet, GoogleNet, ResNet50, MobileNet V2) and classification (VGG16). `NA': `not applicable'.
	\sys speedup the processing time by 1.1$\times$-1.4$\times$  comparing to DeepCache~\cite{xu2018deepcache}, and 1.3$\times$-1.5$\times$ comparing to DeepMon~\cite{huynh2017deepmon}.
	DeepCache's and DeepMon's implementation is not compatible with the two models MobileNet V2 and VGG16, so we are not able to reproduce some results. `NA': `not applicable'.}
	\label{latency0}
\end{figure}

\begin{figure}[t]  
	\centering
	\includegraphics[width=1\linewidth]{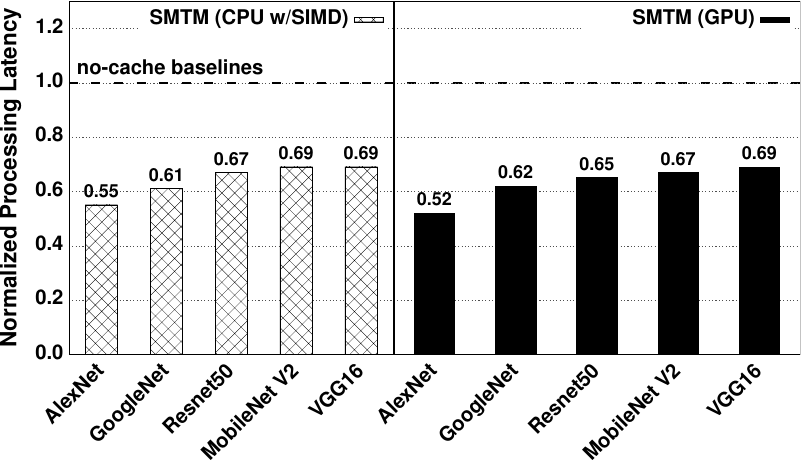}
	\centering
	\caption{Average processing latency of \sys with mobile CPU (w/SIMD) and mobile GPU on action recognition (AlexNet, GoogleNet, ResNet50, MobileNet V2) and classification (VGG16).}
	\label{latency_gpu}
\end{figure}

First, we evaluate the latency reduction on mobile CPU with naive ncnn configuration. 
To make a fair comparison with the state-of-the-arts \cite{xu2018deepcache, huynh2017deepmon}, we set the same ncnn configuration (CPU w/o SIMD) as DeepCache and the results are shown in Figure \ref{latency0}.
The open-sourced implementation of DeepMon \cite{huynh2017deepmon} and DeepCache \cite{xu2018deepcache} are not compatible with the two models MobileNet V2 and VGG16, so we are not able to reproduce some results. Therefore, we only compare the results on AlexNet, GoogleNet, ResNet50. 
It shows that \sys achieves 37.4\% latency reduction on average on three widely used CNN models AlexNet, GoogleNet, ResNet50, while DeepMon and DeepCache have only 10.5\% and 20.9\%. Comparing their performance on different CNN models, \sys reduces the processing time by 1.1$\times$-1.4$\times$ than DeepCache, and 1.3$\times$-1.5$\times$ than DeepMon. For the model AlexNet with only 6 convolutional layers and 3 fully connected layers, \sys can achieve 32.9\% latency reduction, while for the deeper and compact network MobileNet V2, \sys can even save 48.5\% processing time. It shows that \sys achieves better latency reduction on the deeper network, while the performance of DeepMon and DeepCache deteriorates as the models become deeper. 

Then, we test the latency reduction on mobile CPU with SIMD acceleration and GPU. The state-of-the-art method DeepCache requires modifications of the convolution compute kernels, thus it's difficult for it to use the acceleration operation of CPU and GPU.
As \sys directly skips the entire CNN layers and no changes to the CNN forward-path are required, it can be directly applied to any existing deep learning framework and orthogonal to the optimization of the framework itself. 
Figure \ref{latency_gpu} shows the performance of Semantic on mobile CPU with SIMD acceleration and GPU. Compared to no-cache, \sys can have substantial latency reduction, 35.9\% on average on mobile CPU with SIMD acceleration, and 36.8\% on average on mobile GPU. It shows that \sys can work together with various acceleration computation kernels of the inference computing framework. Specifically, for the action recognition scenario (UCF101), the latency reduction brought by \sys range from 31.1\% to 47.7\%, which reveals that our method can efficiently utilize the redundancy in continuous vision. 
On the extreme scenario long-tail CIFAR-100, applying \sys on VGG16 can also achieve 30.6\% latency reduction on mobile GPU and 31.10\% on CPU with SIMD, which shows \sys is robust to scene changes.

\subsection{Accuracy loss}
\label{sec:accuracy}

\begin{figure}[t]  
	\centering
	\includegraphics[width=0.95\linewidth]{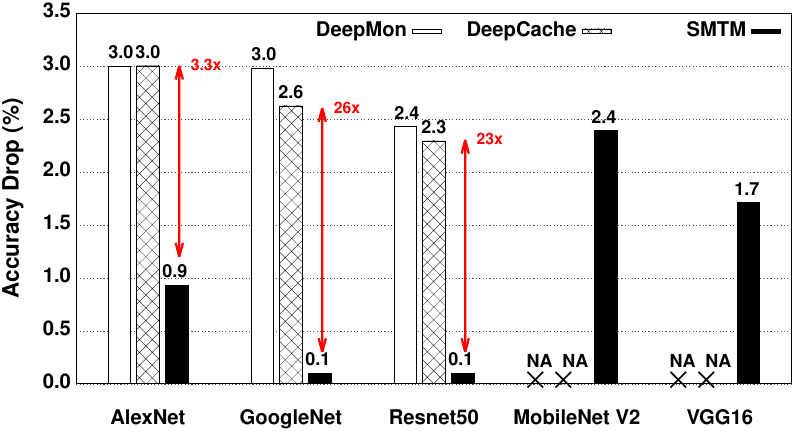}
	\centering
	\caption{Top-1 accuracy drop of \sys on action recognition (AlexNet, GoogleNet, ResNet50, MobileNet V2) and classification (VGG16). `NA': `not applicable'.}
	\label{accuracy}
\end{figure}

We then investigate how much accuracy \sys compromises in return for the latency reduction above. The accuracy drop of \sys is shown in Figure \ref{accuracy}. We can see that on action recognition and image classification scenarios, introducing \sys only leads to 1.05\% latency loss on average on the five CNN models. Compared to DeepMon and DeepCache on UCF101, Semantic achieves much lower accuracy loss on Alexnet and GoogleNet. In detail, the maximum accuracy loss of \sys on the 5 CNN models does not exceed 2.5\%. In particular, for the famous GoogleNet and ResNet50 on UCF101, \sys achieves up to 40.8\% latency reduction with only 0.1\% accuracy loss, which is negligible.
This is because that we have designed a similarity accumulation mechanism that makes the final decision based on the matching on the whole path instead of a single layer, thus minimizing the impact of some layers' wrongly semantic matching for the final decision.

\subsection{Memory overhead}
\label{sec:memory}

\begin{figure}[t]  
	\centering
	\includegraphics[width=0.95\linewidth]{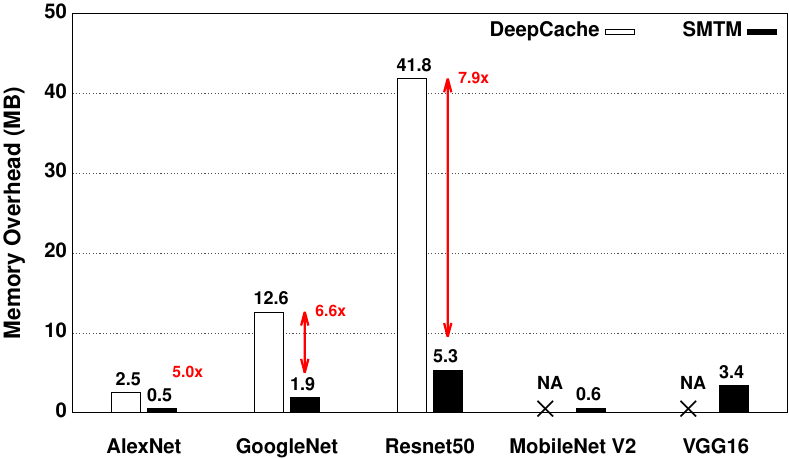}
	\centering
	\caption{The memory overhead of \sys on action recognition (AlexNet, GoogleNet, ResNet50, MobileNet V2) and classification (VGG16). `NA': `not applicable'.}
	\label{memory}
\end{figure}

We also evaluate the memory overhead introduced by \sys. 
The results are shown in Figure \ref{memory}. As observed, the memory overhead occurred by Semantic ranges from \textbf{0.5MB} to \textbf{5.3MB} on the five general CNN models, with an average of 2MB, which is only \textbf{10\%} of the average overhead brought by the state-of-the-art method DeepCache \cite{xu2018deepcache}.
This overhead is quite trivial for the equipped large size of memory in nowadays mobile devices, e.g., 6GB in Google Pixel 4XL.
The reason for the above performance is that, unlike previous methods which store the expensive raw images and feature maps, \sys only need to cache the semantic centers of objects, which are composed of low-dimensional vectors.

\subsection{Energy saving}
\label{sec:energy}

\begin{figure}[t]  
	\centering
	\includegraphics[width=0.95\linewidth]{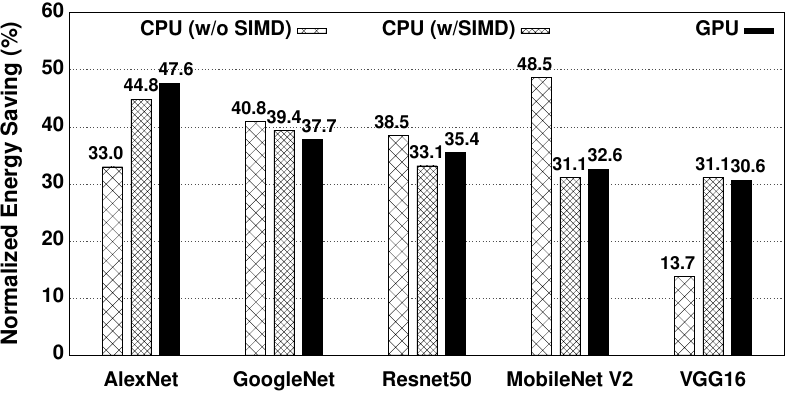}
	\centering
	\caption{The energy saving ratio with different devices on action recognition (AlexNet, GoogleNet, ResNet50, MobileNet V2) and classification (VGG16).}
	\label{energy}
\end{figure}

Next, we investigate the energy consumption of \sys across all the selected test benchmarks. 
The energy consumption is measured via on-device PMIC (power management integrated circuit). The PMIC reports the mobile devices' current and voltage readings at around 800Hz. A single inference is consists of data loading, data prepossessing, inference, and collecting benchmark data. As the inference time is too short, it's difficult to capture enough data to get a valid inference time energy. Therefore, we force the device in an infinite inference loop and measuring the average voltage and current to get the power. Then, we integrate the power with the inference time collected to get the energy reports. Finally, the energy-saving ratio at different devices (CPU and GPU) are shown in Figure \ref{energy}. It shows that \sys can achieve 35.40\% energy saving on average on mobile CPU and 30.60\% on average on mobile GPU. The maximum energy saving on mobile CPU and GPU can be up to 48.55\% and 47.65\%, respectively. 
This saving is mostly from the reduced processing time and reveals that \sys can achieve good performance on different devices.

\subsection{Early exit performance}
\label{sec:exit}

\begin{figure}[t]  
	\centering
	\includegraphics[width=0.95\linewidth]{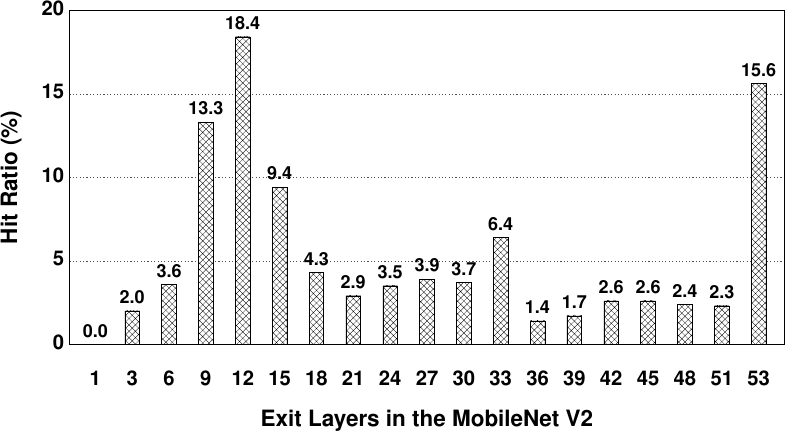}
	\centering
	\caption{Hit ratio at different layers of MobileNet V2 on action recognition. }
	\label{exit_ratio}
\end{figure}

We also report the early exit performance of our \sys across all the selected test benchmarks. Figure \ref{exit_ratio} shows the exit ratio of MobileNet V2 at different layers. During the CNN inference, we first record the exit position of each image and then summarize the exit ratio at each layer for the whole dataset. In Figure \ref{exit_ratio}, the abscissa is the layer position, and the ordinate is the exit ratio. The final bar in the figure is the ratio of full inference. We perform semantics matching after each Inception module in MobileNet V2.
Our experiments show that the exit ratio varies from layer to layer, which demonstrates that the semantic features of these middle layers have different characteristics. For the four CNN models on action recognition scenario, the exit ratio at all layers excepts the last one on the dataset ranges from \textbf{67.4\%} to \textbf{92.7\%}, and more than 50\% of images on average can exit the inference in the first half of the network. For the extreme scenario image classification with rapid scene changes, there are also 63.8\% images that can early exit the inference.
The results indicate \sys can make good use of the temporal redundancy in mobile videos to accelerate the CNN inference and well adapt to the various scenarios.

\subsection{Choice of parameter}
\label{parameter}

\begin{figure}[t]  
	\centering
	\includegraphics[width=0.95\linewidth]{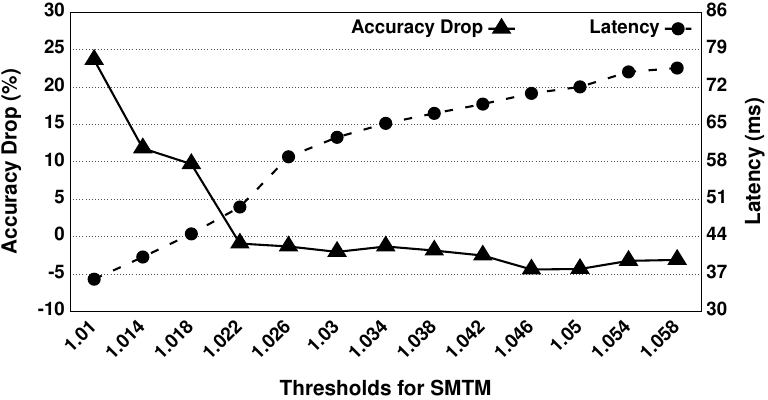}
	\centering
	\caption{The accuracy drop and averaging processing latency on CPU (w/SIMD) with different thresholds with GoogleNet on action recognition. It shows that, as the threshold changes, there is a trade-off between accuracy and latency. }
	\label{trade-off}
\end{figure}

In our \sys framework, the variable confidence threshold $\tau$ for early exit can be used to make the trade-off between the latency reduction and accuracy loss. The threshold $\tau$ is the key to decide whether the feature of the new input can be matched with the cached features of the selected objects in the fast memory. The results in Figure \ref{trade-off} show how $\tau$ can affect the processing latency and accuracy (GoogleNet on UCF101). As expected, lower $\tau$ will bring higher latency reduction, thus leading to more accuracy loss. Besides, we noticed that when a larger threshold is set, \sys can even achieve an accuracy performance which slight over the baseline, which shows that there are some cases where the baseline wrongly classifies the images while our \sys does it correctly. 
This is because that we have designed adaptive semantic centers that not only encoding the semantics of objects from the training dataset but also the objects from the test scenario, thus improving the representation ability of the semantic centers in the memory.
Besides, the threshold can also be set by application developers to adapt to task-specific requirements. For applications that are not very sensitive to the output accuracy, developers can aggressively use a smaller $\tau$ to achieve higher latency reduction.

\subsection{The effect of adaptive memory}
\label{Eva:adaptive}

\begin{table}[t]
    \setlength{\abovecaptionskip}{0.cm}
	\begin{center}
		\caption{The impact of adaptive cache size. Tested on ResNet50 model. }
		\label{tab:cache_size}
		\begin{tabular}{ccc}
			\noalign{\smallskip}
			\hline
			 & Hit ratio & Latency reduction  \\ \hline
			\sys (Constant)  & 65.39\%  &  25.21\%     \\ 
			\sys (Adaptive)  & 87.00\%  & 38.46\%   \\ \hline
		\end{tabular}
	\end{center}
\end{table}

\begin{figure}[t]  
	\centering
	\includegraphics[width=0.9\linewidth]{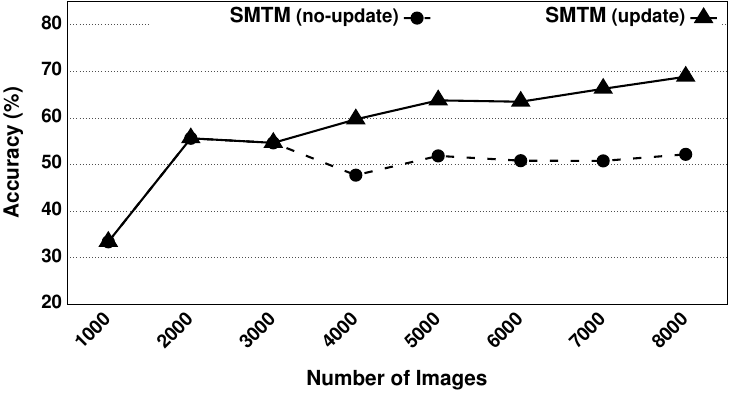}
	\centering
	\caption{The impact of adaptive semantics center on the prediction accuracy on ResNet50. }
	\label{adaptive_centers}
\end{figure}

In this section, we evaluate the effect of the proposed two techniques in Section \ref{sec:EEM} to overcome the high dynamic scenarios in the mobile vision task. 

First, we evaluate the effect of adaptive cache size in fast memory. We set a constant cache size (size=5) in the fast memory as `\sys (Constant)' and set this constant number as the initial cache size in `\sys (Adaptive)'. The other settings remain the same. The results in Table \ref{tab:cache_size} shows that compared to `\sys (Constant)', `\sys (Adaptive)' increases the hit ratio by 21.6\%, which brings 1.5$\times$ latency reduction.

There is a trade-off between accuracy and latency when varying the fast memory size. Based on our experiments, we can indeed found an optimal fast memory size with the best trade-off in every small period of time. However, due to the high dynamicity in mobile data, the optimal size continues to change throughout the long video. In real-world applications, the optimal size is not known in advance. Therefore, we adopt probability estimation to predict a memory size in real-time in the paper. The predicted memory sizes are usually sub-optimal, but the overall performance exceeds the constant size by a large margin. In addition, the constant size in Table \ref{tab:cache_size} is already the optimal size we found before, which can contribute to better performance on the entire test dataset.

Then, we evaluate the effect of adaptive semantic centers. The results in Figure \ref{adaptive_centers} show that compared to the baseline (no-update), updating the semantic center adaptively can gradually improve the prediction accuracy and finally achieves 16.9\% accuracy improvement on action recognition.

The above two experiment shows the proposed two techniques are beneficial to the \sys adapt to the scenario changes in mobile vision tasks.

\section{Discussion}

This section highlights some of the limitations of \sys and discusses possible future research directions.

\textbf{Generalized to diverse vision tasks.} Although \sys currently focuses on recognition and classification tasks, we believe that the proposed memory design can be applied to many applications.
First, recognition and classification are two general mobile vision scenarios, which include many applications involving deep learning in mobile/wearable devices. Second, for many multi-stage mobile vision tasks (such as object detection, etc), they usually also need to do recognition and classification. Thus, by leveraging the temporal redundancy in mobile videos, \sys can also be potentially generalized to improve the processing in these tasks.

\textbf{Beyond vision tasks.} While \sys currently focuses on continuous vision tasks, its key design of semantic memory can be potentially generalized beyond to other types of ML tasks such as natural language processing and speech recognition.
This is because the temporal redundancy universally exists in those tasks, e.g., hot-spot keywords in the input method and voice assistants.

\section{Related Work}

\textbf{CNN Cache.} 
A few recent works~\cite{huynh2017deepmon,xu2018deepcache,cavigelli2017cbinfer} also exploit the temporal redundancy to accelerate continuous visual tasks.
They match the similar blocks (or pixels) of images or feature maps and reuse the intermediate partial results to skip computations.
Those methods need to cache high-dimensional video frames and feature maps of CNN and then perform an expensive lookup on them.
\sys fundamentally differs from them in
(1) \sys encodes feature maps into low-dimensional semantic vectors of each object and executes distance measure between these feature vectors.
(2) \sys directly exits the inference when an object is matched.
FoggyCache \cite{guo2018foggycache} focuses on cross-device cache reuse, which is orthogonal to \sys.
EVA$^{2}$ \cite{buckler2018eva2} proposes a cache extension to CNN accelerator, which is not compatible with mobile scenarios.
By contrast, \sys is designed and implemented to run on general-purpose processors which are widely available on commodity mobile/wearable devices.

\textbf{Multi-branch neural architectures.} Some efforts propose multi-branch architecture into neural network designs to accelerate the CNN inference \cite{figurnov2017spatially, li2017not, wu2020emo, wang2018skipnet, laskaridis2020spinn}. For example, BranchyNet~\cite{teerapittayanon2016branchynet} trains a network that allows simple samples to exit inference at the early layers. Similarly, SkipNet \cite{wang2018skipnet} allows easy samples to skip some middle layers during the inference and BlockDrop \cite{wu2018blockdrop} skips some middle blocks for easy samples.
\textbf{While \sys also early exits at different layers, its rationale is fundamentally different from the above-mentioned multi-branch networks: Our \sys leverages opportunity from temporal-dimension, where inferences exit when repeated objects are observed from recent memory. Our key insight is orthogonal to theirs.} 
SPINN \cite{laskaridis2020spinn} proposes a distributed inference system that employs synergistic device-cloud computation and progressive inference to accelerate CNN inference. However, putting the input to the cloud will bring significant privacy concerns. By contrast, \sys is proposed to benefit directly on-device CNN inference.
Furthermore, these above methods usually need to retrain the CNN models, which is a quite time-consuming process.
\sys is compatible with commodity, pre-trained CNN models, requiring zero effort from developers.

\textbf{On-device CNN optimizations.}
Besides the aforementioned methods, extensive efforts have been made to optimize the CNN inference so they can be affordable on mobile/wearable devices, such as weight quantization \cite{kim2019mulayer, courbariaux2016binarized, cao2019seernet}, pruning \cite{niu2020patdnn, han2015deep, liu2019metapruning, li2020weight}, hardware-based acceleration \cite{zhang2015optimizing, zhang2018caffeine, xu2020approximate}, model compression for mobile devices \cite{mathur2017deepeye, yi2020heimdall, yeo2020nemo, lee2020fast, zhang2020mdldroidlite}, etc.
To our knowledge, \sys is the first system that accelerates continuous mobile vision by infusing the priming effect mechanism with CNN inference and is orthogonal to existing model-level or hardware-level optimizations.

\section{Conclusions}

In this paper, we propose a novel memory mechanism, called semantic memory, to speed up on-device CNN inference.
The design of our memory mechanism is based on the observation of high temporal redundancy of continuous visual input on mobile scenarios and how human brains perform fast recognition of repeatedly presented objects.
Extensive experiments show the superior performance of our system against existing cache systems.

\begin{acks}
Mengwei Xu was supported in part by the Fundamental Research Funds for the Central Universities and National Natural Science Foundation of China under grant numbers 62032003, 61922017, and 61921003.
\end{acks}

\bibliographystyle{ACM-Reference-Format}
\bibliography{egbib}

\end{document}